\documentclass{article}
\usepackage{ijcai21}

\usepackage{times}
\usepackage{soul}
\usepackage{url}
\usepackage[hidelinks]{hyperref}
\usepackage[utf8]{inputenc}
\usepackage[small]{caption}
\usepackage{graphicx}
\usepackage{amsmath}
\usepackage{amsthm}
\usepackage{booktabs}
\usepackage{helvet} 
\usepackage{courier} 
\usepackage{epstopdf}
\usepackage{tabularx}
\usepackage{multirow}
\usepackage{subcaption}
\usepackage{color}
\usepackage{multirow}
\usepackage{comment}
\usepackage{pythontex}
\usepackage[ruled, linesnumbered]{algorithm2e}

\title{EventDrop: Data Augmentation for Event-based Learning}

\author{
Fuqiang Gu$^1$
\and
Weicong Sng$^2$\and
Xuke Hu$^{3}$\And
Fangwen Yu$^4$\thanks{Corresponding Author}
\affiliations
$^1$ College of Computer Science, Chongqing University, China\\
$^2$ School of Computing, National University of Singapore, Singapore\\
$^3$ Institute of Data Science, German Aerospace Center, Germany\\
$^4$ Department of Precision Instrument, Tsinghua University, China
\emails
gufq@cqu.edu.cn,
sngweicong@comp.nus.edu.sg,
xuke.hu@dlr.de,
yufangwen@tsinghua.edu.cn
}

\begin{document}

\maketitle

\begin{abstract}
	The advantages of event-sensing over conventional sensors (e.g., higher dynamic range, lower time latency, and lower power consumption) have spurred research into machine learning for event data. Unsurprisingly, deep learning has emerged as a competitive methodology for learning with event sensors; in typical setups, discrete and asynchronous events are first converted into frame-like tensors on which standard deep networks can be applied. However, over-fitting remains a challenge, particularly since event datasets remain small relative to conventional datasets (e.g., ImageNet). In this paper, we introduce EventDrop, a new method for augmenting asynchronous event data to improve the generalization of deep models. By dropping events selected with 
	various strategies, we are able to increase the diversity of training data (e.g., to simulate various levels of occlusion). From a practical perspective, EventDrop is simple to implement and computationally low-cost. Experiments on two event datasets (N-Caltech101 and N-Cars) demonstrate that EventDrop can significantly improve the generalization performance across a variety of deep networks.  
	
\end{abstract}

\section{Introduction}
\begin{figure}[htb]
	\centering
	\includegraphics[width=\linewidth]{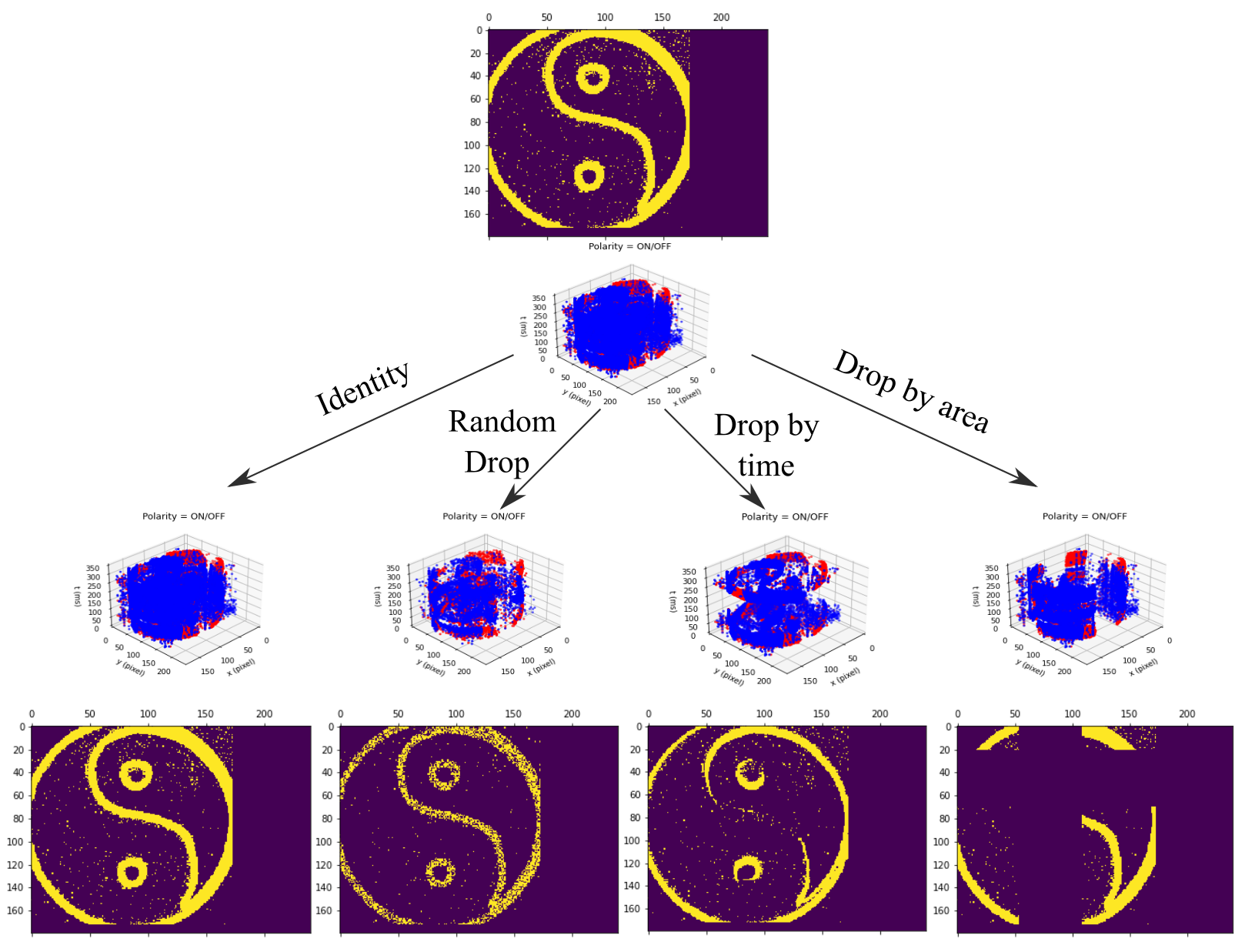}
	\caption{An example of augmented events with EventDrop. For better visualization, the event frame representation is used to visualize the outcome of augmented events. }
	\label{fig:intro_eventdrop}
\end{figure}
Event sensors, such as DVS event cameras \cite{patrick2008128x} and NeuTouch tactile sensor \cite{taunyazov2020event}, are bio-inspired devices that mimic the efficient event-driven communication mechanisms of the brain. Compared to conventional sensors (e.g., RGB cameras), which synchronously capture the scene at a fixed rate, event sensors asynchronously report the changes (called events) of the scene. For example, DVS cameras capture the changes in luminosity over time for each pixel independently rather than intensity images as RGB cameras do. Event sensors usually have the advantages of higher dynamic range, higher temporal resolution, lower time latency, and higher power efficiency \cite{gehrig2019end}. These advantages have stimulated research into machine learning for event data. Unsurprisingly, deep learning, which performs extremely well on a variety of tasks, remains a competitive method for learning with event sensors.

A challenging problem in deep learning is over-fitting, which causes a model that exhibits excellent performance on training data to degrade dramatically when validated against new and unseen data. A simple solution to the over-fitting problem is to significantly increase the amount of labeled data, which is theoretically feasible but may be cost-prohibitive in practice. The over-fitting problem is more severe in learning with event data since event datasets remain small relative to conventional datasets (e.g., ImageNet). 

Data augmentation is a way to increase both the amount and diversity of data from existing data, which can improve the generalization ability of deep learning models. For images, common augmentation techniques include Translation, Rotating, Flipping, Cropping, Contrast, Sharpness, Shearing, etc \cite{cubuk2019autoaugment}. Event data are fundamentally different from frame-like data (e.g., images), and hence we cannot directly use these augmentation techniques that are originally developed for frame-like data to augment asynchronous event data. 

In this paper, we present EventDrop, a novel method to augment event data by dropping events. This is motivated by the observations that the number of events significantly changes over time even for the same scene and that occlusion often occurs in many vision tasks. To address these issues, we propose three strategies to select certain events to be dropped, including \textit{Random drop}, \textit{Drop by time}, and \textit{Drop by area}. Figure \ref{fig:intro_eventdrop} shows a specific example of augmented events with different operations of EventDrop. Through these augmentation operations, we can enlarge both the amount of training data as well as the data diversity, which will benefit deep learning models. To the best of our knowledge, EventDrop is the first work to augment asynchronous event data by dropping events.

The closely-related works to this study are Dropout \cite{srivastava2014dropout}, Cutout \cite{devries2017improved}, RE \cite{zhong2020random} and SpecAugment \cite{park2019specaugment}, all of which introduce some noise to improve the generalization ability of deep learning models. However, Dropout drops the units and their connections in the models that can be in the intermediate layers, while our method only drops events in the input space. EventDrop can be considered as an extension of Dropout in the input space. Compared to Cutout and RE, both of which deal with images by considering occlusion, EventDrop works with event data and deals with both sensor noise and occlusion. SpecAugment works on audio, while our method deals with event-based data.

In summary, our main contributions are as follows:
\begin{itemize}
	\item We propose EventDrop, a novel method for augmenting asynchronous event data, which is simple to implement, computationally low-cost, and can be applied to various event-based tasks.
	\item We evaluate the proposed method on two public event datasets with different event representations. Experimental results show that the proposed method significantly improves the generalization performance across a variety of deep networks.  
\end{itemize}

\section{Related Work}
\subsection{Event-based Learning}
Event-based learning has been increasingly popular due to the advantages of event sensors (e.g, low time latency, low power consumption, and high dynamic range) \cite{gallego2019event,lee2020enabling}. Event-based learning algorithms can be grouped into two major approaches. One approach is to first convert asynchronous events into frame-like data, such that frame-based learning methods can be applied directly (e.g., state-of-the-art DNNs). Some representative works include event frame \cite{rebecq2017real}, Event Count Image \cite{maqueda2018event}, Voxel Grid \cite{zhu2019unsupervised}, and Event Spike Tensor (EST) \cite{gehrig2019end}. While such methods can make use of the powerful ability of modern DNNs through event conversion, they may discard some useful information about the events (e.g., polarity, temporal information, density).

The other approach is to directly use spiking neural networks (SNNs) on the asynchronous event data. The event-driven property of SNNs makes them inherently suitable for dealing with event data. Compared to standard DNNs, SNNs are more biologically plausible and more energy efficient when implemented on neuromorphic processors. Event-based learning with SNNs has been used for object recognition \cite{gu2020tactilesgnet}, visual-tactile perception  \cite{taunyazov2020event}, etc. While SNNs are attractive for dealing with event data, the spike function is not differentiable and hence one cannot directly use backpropagation methods to train the SNNs. Several solutions have been proposed to address this issue, such as converting DNNs to SNNs and approximating the derivative of the spike function~\cite{wu2019direct}. However, the overall performance of SNNs is often inferior to standard DNNs.  

In this study, we focus on data augmentation for event-based learning with DNNs, but the proposed method can be also applied for SNN-based methods. 

\subsection{Regularization}
Regularization is a key technique for mitigating over-fitting in the training of deep learning models. Common regularization strategies include weight decay, and Dropout \cite{goodfellow2016deep}. The basic idea of weight decay is to penalize the model weights by adding a term to the loss function. Popular forms of weight decay are $L^1$ and $L^2$ regularization.
Dropout is also a widely-used regularization technique, which simulates sparsity for the layer it is applied to. In the standard Dropout method \cite{srivastava2014dropout}, units and their connections are randomly dropped out from the model with a certain probability (e.g., 0.5) during training. Many variants have been proposed to further improve the speed or regularization effectiveness \cite{labach2019survey}. Compared to Dropout, this study drops events in the input space rather than drops the units and their connections in the models. 
\subsection{Data Augmentation}
Data augmentation can be also regarded as a regularization method that improves the generalization ability of deep learning models.
It is widely accepted that deep learning models over-fit, and benefit strongly from larger datasets. Data augmentation is a practical technique to increase the amount of training data as well as the data diversity.  Many studies have demonstrated that deep learning models can significantly improve their generalization ability by applying some transforms on the input images \cite{krizhevsky2012imagenet}, such as Translation, Rotation, Flipping, and Cropping. Recently, A popular augmentation technique called SamplePair \cite{inoue2018data} is proposed for image classification, which creates a new sample from one image by overlaying another image that is randomly selected from training data.

Different from existing works that deal with images, this study works on the augmentation of event data, which remains unexplored to the best of our knowledge. We focus particularly on the object occlusion problem and noisy event data. The closely-related works that also deal with occlusion are Cutout \cite{devries2017improved}, RE \cite{zhong2020random}, and SpecAugment \cite{park2019specaugment}. Specifically, Cutout applies a fixed-size zero mask to a random location of each input image, while RE erases the pixels in the randomly selected region with random values. Compared to Cutout and RE, which deal with images, our approach deals with event data and considers both object occlusion and sensor noise. SpecAugment is an augmentation method for audio, which operates on the log mel spectrogram of the input audio. Both SpecAugment and our method are inspired by Cutout. However, SpecAugment works on audio, while our method deals with event data (which are asynchronous events).

\section{Event Representation}
\label{sec:event_representation}

\begin{figure}[ht]
	\centering
	\includegraphics[width=0.9\linewidth]{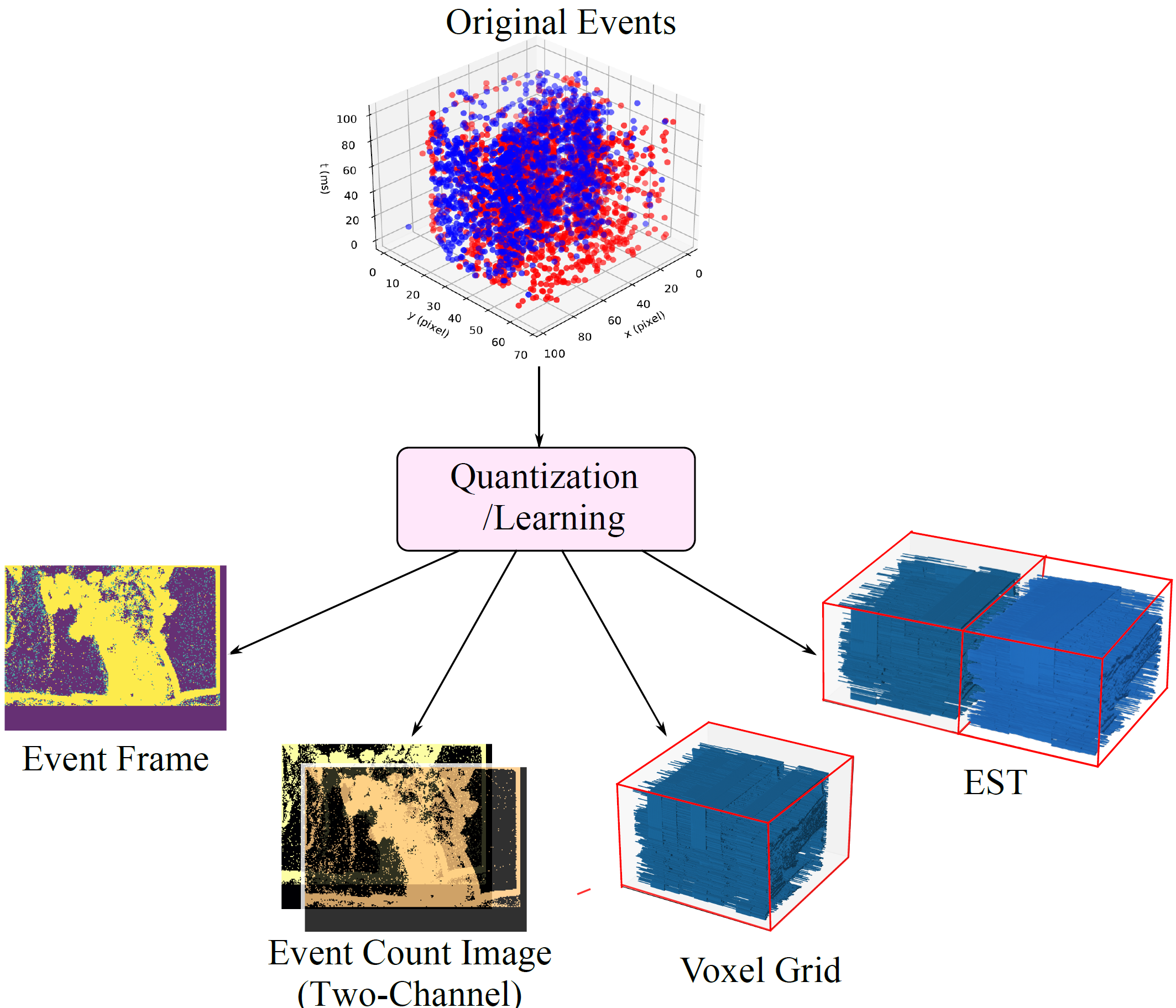} 
	\caption{General framework of converting events into popular event representations. The original asynchronous events can be transformed into frame-like data through quantization or learning (e.g., neural networks).  }
	\label{fig:event_representation}
\end{figure}

State-of-the-art DNNs usually deal with frame-like data (e.g., images, videos) and cannot be directly used for event data since event data are a stream of asynchronous events. An individual event alone, contains little information about the scene. To make use of event data, certain methods have been proposed to learn useful frame-like representations that can be exploited by DNNs. Popular event representations include Event Frame \cite{rebecq2017real}, Event Count Image \cite{maqueda2018event}, Voxel Grid \cite{zhu2019unsupervised}, and EST \cite{gehrig2019end}, which we will introduce in the following. Figure \ref{fig:event_representation} shows the general framework of converting asynchronous event data into popular event representations.

Let $\mathbf{\varepsilon}$ be a sequence of events, which encode the location, time, polarity (sign) of the changes. It can be described as:
\begin{equation}
\mathbf{\varepsilon} = \{e_i \}_{i=1}^I = \{x_i, y_i, t_i, p_i \}_{i=1}^I,
\end{equation}
where $(x_i, y_i)$ is the coordinate of the pixel triggering the event $e_i$, $t_i$ is the timestamp when the event is generated, and $p_i$ is the polarity of the event. The polarity takes two values: $1$ and $-1$, representing positive and negative events, respectively. $I$ is the number of events. 

Event Frame represents events using the histograms of events for each pixel, which can be written as (denoted by $V_{EF}$):
\begin{equation}
V_{EF}(x_l, y_m) = \sum_{e_i \in \mathbf{\varepsilon} } \delta(x_l-x_i) \delta(y_m-y_i),
\end{equation}
\begin{equation}
\delta(a) = 
\begin{cases}
1, \ \text{if} \ a = 0 \\
0, \ \text{otherwise},
\end{cases}
\end{equation}
where $\delta(\cdot)$ is an indicator function. $(x_l, y_m)$ is the pixel coordinate in the Event Frame representation, and $x_l \in \{0, 1, \cdots, W-1\}$, $y_m \in \{0, 1, \cdots, H-1 \}$. The Event Frame can be regarded as a 2D image with a resolution of $H\times W$. 

Event Count Image is similar to Event Frame, but it uses separate histograms for positive events and negative events. Event Count Image $V_{EC}$ is described as:
\begin{equation}
V_{EC}(x_l, y_m, \pm) = \sum_{e_i \in \mathbf{\varepsilon}_{\pm} } \delta(x_l-x_i) \delta(y_m-y_i),
\end{equation}
where $\mathbf{\varepsilon}_{+}$ and $\mathbf{\varepsilon}_{-}$ are event sequences with positive polarity and negative polarity, respectively. The Event Count Image can be regarded as a two-channel image with each channel corresponding to one polarity.

Voxel Grid $V_{VG}$ considers the temporal information of the events, which is not explicitly handled in Event Frame and Event Count Image. It is written as
\begin{equation}
V_{VG}(x_l, y_m, c_n) = \sum_{t_{n-1} < t_i \leq t_n} \delta(x_l-x_i) \delta(y_m-y_i)  \mathbf{1}_{t_i},
\end{equation}
\begin{equation}
t_n = t_1 + (c_n + 1) \Delta T,
\end{equation}
where $\mathbf{1}_{t_i}$ is an indicator function, which takes 1 when $t_i$ is in the interval $(t_{n-1}, t_n]$ and 0 otherwise. $c_n$ is the temporal index of the Voxel Grid representation, and $c_n \in \{0, 1, \cdots, C-1 \}$. $\Delta T$ is temporal bin size. 

Similar to Voxel Grid, EST is also a grid-based representation that is learned end-to-end directly from asynchronous event data through differentiable kernel convolution and quantization. EST considers both temporal information and polarity about events, which is described as: 
\begin{equation}
\begin{split}
& V_{EST}(x_l, y_m, c_n, \pm) \\ & =  \sum_{e_i \in \mathbf{\varepsilon}_{\pm}} f_{\pm}(x_i, y_i, t_i) k(x_l-x_i, y_m-y_i, t_n - t_i),
\end{split}
\end{equation}
where $f_{\pm}(x, y, t)$ is the normalized timestamp, and $f_{\pm}(x, y, t)=\frac{t-t_1}{\Delta T}$ where $t_1$ is the first timestamp, $\Delta T$ is the bin size. $k(x, y, t)$ is a trilinear kernel, which is written as 
\begin{equation}
k(x, y, t) = \delta (x, y) \text{max} (0, 1- \left |\frac{t}{\Delta T}\right|).
\end{equation}

In addition to the above representations, there are other event representations such as HOTS \cite{lagorce2016hots}, HATS \cite{sironi2018hats}, and Matrix-LSTM \cite{cannici2020matrix}. In this study, we take the four representations as representatives to analyze how EventDrop enhances the performance of DNNs. 

\begin{figure}[ht]
	\centering
	\includegraphics[width=\linewidth]{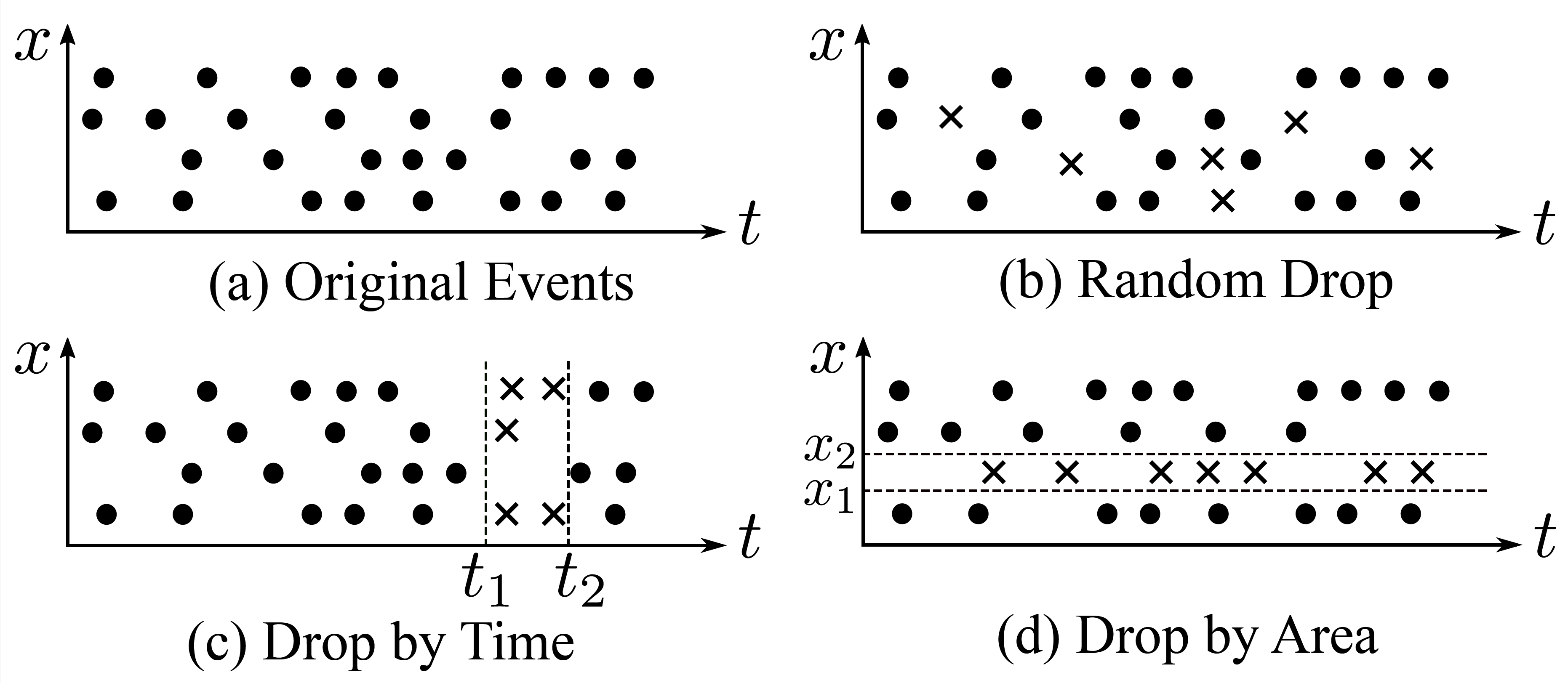}
	\caption{Strategies used by EventDrop, where $t$ indicates time dimension, $x$ denotes the pixel coordinate (only one dimension is shown here for clarity). Dots represent original events, and $\times$ denotes the events to be dropped. Dashed lines represent threshold borders. (a) Original events that are triggered asynchronously. (b) \textit{Random drop} strategy. (c) \textit{Drop by time} strategy. (d) \textit{Drop by area} strategy. }
	\label{fig:drop_strategy}
\end{figure}

\section{Proposed Method: Augmenting Event Data by Dropping Events}
\subsection{Motivation}
This work is motivated by two observations. The first observation is that the output of event cameras for the same scene under the same lighting condition may vary significantly over time. This may be because event cameras are somehow noisy, and events are usually triggered when the change about the scene reaches or surpasses a threshold. By randomly dropping a proportion of events, it is possible to improve the diversity of event data and hence increase the performance of downstream applications.  

The second observation is that occlusion often occurs in many tasks such as object recognition and tracking. The generalization ability of a machine learning model depends highly on the diversity of training data, including various levels of occlusion. However, the available training data usually suffer from limited variance in the occlusion level. A machine learning model trained on the data with limited or no (totally visible) occlusion variance may generalize poorly on new samples that are partially occluded. By generating new samples that simulate partially occluded cases, the model is able to better recognize objects with partial occlusion.

\subsection{Strategies of Dropping Events}
To address the above issues, we propose three strategies of dropping events to augment event data, namely \textit{Random drop}, \textit{Drop by time}, and \textit{Drop by area}. The first strategy is to prepare the model for noisy event data. The other two strategies are for simulating the occlusion problem. Figure \ref{fig:drop_strategy} illustrates the idea of different dropping strategies. We describe the three strategies in the following. 
\begin{itemize}
	\item \textbf{Random drop}. The basic idea of \textit{Random drop} is to randomly drop a proportion of events in the sequence. This is to overcome the noise from the event sensors. 
	
	\item \textbf{Drop by time}. \textit{Drop by time} is to drop events triggered within a random period of time. It tries to increase the diversity of training data by stimulating the case that objects are partially occluded during certain time period.
	
	\item \textbf{Drop by area}. \textit{Drop by area} is to drop events triggered within a randomly selected pixel area. It also aims to improve the data diversity by simulating various cases that some parts of the objects are partially occluded. 
	
\end{itemize}

\subsection{Implementation}
\label{sec:implementation}

\begin{algorithm}[!htb]
	\caption{Procedures of augmenting event data with EventDrop}
	\label{alg:event_drop}
	\SetKwInOut{Input}{Input}
	\SetKwInOut{Output}{Output}
	\Input{A sequence of events $\mathbf{\varepsilon}= \{e_i \}_{i=1}^I = \{x_i, y_i, t_i, p_i \}_{i=1}^I$, pixel resolution ($W$, $H$).} 
	\Output{Augmented event sequences $\mathbf{\varepsilon}^*$.}
	Initialize the $\mathbf{\varepsilon}^*$ to an empty set, namely $\mathbf{\varepsilon}^* = \{\}$; \\
	$Operation$ $\gets$ Random.choice($identity$, $drop\_by\_time$, $drop\_by\_area$, $random\_drop$); \\
	\If{$Operation == identity$}{
		$\mathbf{\varepsilon}^* \gets \mathbf{\varepsilon}$; \\
	}
	\If{$Operation == drop\_by\_time$}{
		$\rho \gets Rand(1, 10)/10; $ \\
		$T_{min} \gets Rand(t_1, t_I)$;\\
		$T_{max} \gets max(t_I, T_{min} + \rho*(t_I - t_1))$;\\
		\For{$e_i \in \mathbf{\varepsilon}$ }{
			\If{$(t_i < T_{min})\| (t_i > T_{max})$}{
				Add $e_i$ into $\mathbf{\varepsilon}^* $; 
			}
		}	
	}
	\If{$Operation == drop\_by\_area$}{
		$\rho \gets Rand(1, 6)/20; $ \\ 
		$x_0 \gets Rand(0, W)$; \\
		$y_0 \gets Rand(0, H)$; \\
		\For{$e_i \in \mathbf{\varepsilon}$}{
			\If{	$(x_i \in [x_0, x_0 + \rho*W]) \& (y_i \in [y_0, y_0 + \rho*H]$}{
				Do nothing; 
			}
			\Else{
				Add $e_i$ into $\mathbf{\varepsilon}^* $; 
			}
		}
	}
	\If{$Operation == random\_drop$}{
		$\rho \gets Rand(1, 10)/10; $ \\ 
		$\mathbf{\varepsilon}^* = Random.choices(\mathbf{\varepsilon}, I*(1-\rho))$

	}
	\Return $\mathbf{\varepsilon}^*$.		
\end{algorithm}

In this section, we describe the implementation of EventDrop. Algorithm \ref{alg:event_drop} gives the procedures of augmenting event data with EventDrop. This algorithm takes as input a sequence of asynchronous events and corresponding image resolution $(W, H)$. We first define four augmentation techniques, namely \textit{Identity}, \textit{Random drop}, \textit{Drop by time}, and \textit{Drop by area}, and conduct one augmentation technique that is randomly selected on the event sequence. The random policy exploration in \cite{cubuk2020randaugment} is adopted in this study due to its simplicity and excellent performance. The probability $p$ of each of these augmentation operations being chosen is set to equal (namely, $ p=0.25$). The magnitude of \textit{Random drop} and \textit{Drop by time} is discretized into 9 different levels and that of \textit{Drop by area} into 5 levels. Specifically, when conducting the \textit{Drop by time} operation, a magnitude is first randomly selected and then a period of time is selected. The events triggered within the selected time period would be deleted from the event sequence, the remaining event sequence will be returned as the output of the algorithm. In the \textit{Drop by area} operation, a pixel region is first selected by a random magnitude and a random location, and then the events within the selected region would be dropped. 
In the \textit{Random drop} operation, a proportion of events are randomly selected to be dropped. Overall, EventDrop is simple to implement and computationally low-cost. We have implemented EventDrop in PyTorch and the source code is available at 	\href{https://github.com/fuqianggu/EventDrop}{  \textit{https://github.com/fuqianggu/EventDrop}}.

\section{Experiments and Results}
\subsection{Datasets}
We evaluate the proposed EventDrop augmentation technique using two public event datasets: N-Caltech101 \cite{orchard2015converting} and N-Cars \cite{sironi2018hats}. N-Caltech101 (Neuromorphic-Caltech101) is the event version of the popular Caltech101 dataset \cite{fei2004learning}. To convert the images to event sequences, an ATIS event camera was installed on a motorized pan-tilt unit, and automatically moved while pointing at images from the original dataset (Caltech101) that were shown on a LCD monitor. N-Cars (Neuromorphic-Cars) is a real-world event dataset for recognizing whether a car is present in a scene. It was recorded using an ATIS camera that was mounted behind the windshield of a car. 

\subsection{Experiment Setup}
We evaluate the proposed method using four state-of-the-art deep learning architectures, namely ResNet-34 architecture \cite{he2016deep}, VGG-19 \cite{simonyan2014very}, MobileNet-V2 \cite{sandler2018mobilenetv2}, and Inception-V3 \cite{szegedy2016rethinking}. All the networks are pretrained on ImageNet \cite{russakovsky2015imagenet}. Since the number of input channels and output classes for our case are different from these pre-trained models, we adopt the approach used in \cite{gehrig2019end} and replace the first and last layer of the pre-trained models with random weights, and then fine-tune all the parameters on the task. 

Since event data are a stream of asynchronous events and cannot be directly applied to deep nets, we consider and implement the four event representations introduced in Section 3. For the implementation of EST, we replace the neural network with a trilinear kernel to convolve with the normalized timestamps for computational efficiency. 
Note that the considered deep learning models take as input 2D images, while some event representations we considered (e.g., Voxel Grid and EST) are 3D or 4D tensor. To adapt to these pretrained model, we concatenate the event representation along the polarity and/or temporal dimension as channels.

The Adam optimizer is used to train the model by minimizing the cross-entropy loss. The initial learning rate is set to $1\times 10^{-4}$ until the iteration reaches up to 100, after which the learning rate is reduced by a factor of 0.5 every 10 iterations. The total number of iterations is set to 200. We use a batch size of 4 for both datasets. To conduct a robust evaluation, we run the model on each dataset for multiple rounds with different random seeds, and report the mean and standard deviation values. We perform early stopping on a validation set using the splits provided by the EST \cite{gehrig2019end} on N-Caltech101 and 20\% of the training data on N-Cars. 

\subsection{Results on N-Caltech101}
We first analyze the results of EventDrop on the N-Caltech101 dataset. The results from the same models without data augmentation are considered as the baselines. Table \ref{tab:results_ncaltech101} compares the performance of EventDrop and the baselines. We can see that EventDrop improves the performance of all the models used with different event representations. The accuracy achieved with Voxel Grid and EST representations is much higher than that with Event Frame and Event Count representations. This is attributed to the fact that Voxel Grid and EST contain temporal information about the events that is discarded by Event Frame and Event Count. Since EST further considers the polarity  information about the events, it behaves slightly better than Voxel Grid. The same trend can be found when comparing Event Frame (without polarity information) and Event Count (with polarity information). Among these deep nets, MobileNet-V2 seems to perform slightly better than ResNet-34 and Inception-V3, while VGG-19 performs the worst, probably because it is relatively old. 

\begin{table}  	
	\centering
	\resizebox{\linewidth}{!}{
		\begin{tabular}{cccccccc}
			\toprule[0.5pt]
			\multirow{2}{*}{\bf{Model}}  & \multirow{2}{*}{\bf{Representation}} &  \multicolumn{2}{c}{\textbf{Average Accuracy (Std)}} \\ \cline{3-4}
			& & \textbf{Baseline} & \textbf{EventDrop}  \\ 
			\midrule[0.5pt]
			\multirow{4}{*}{ResNet-34} 
			& Event Frame & 77.39 (0.78) & 78.20 (0.15) \\
			& Event Count & 77.75 (0.64)  & 78.30 (0.29)\\
			& Voxel Grid & 82.47 (0.80) & 82.57 (0.42)\\
			& EST & 83.91 (0.44) & \textbf{85.15 (0.36)}\\ \cline{1-4}
			
			\multirow{4}{*}{VGG-19} 
			& Event Frame &  72.31 (1.38) & 74.99 (0.67) \\
			& Event Count &  73.02 (1.05) & 75.01 (0.57)\\
			& Voxel Grid    &  76.63 (0.81) & 77.28 (0.45)\\
			& EST         &  78.88 (0.79)& \textbf{79.55 (1.25)}\\ \cline{1-4}
			
			\multirow{4}{*}{MobileNet-V2} 
			& Event Frame & 79.08 (0.84)& 82.19 (0.63)\\
			& Event Count & 79.68 (1.09)& 82.31 (0.72)\\
			& Voxel Grid    & 83.12 (0.55)& 85.56 (0.79)\\
			& EST         & 84.76 (0.64)& \textbf{87.14 (0.54)}\\ \cline{1-4}
			
			\multirow{4}{*}{Inception-V3}
			& Event Frame & 80.01 (0.81) & 81.46 (0.55)\\
			& Event Count & 80.15 (0.56) & 81.01 (0.81)\\
			& Voxel Grid    & 82.68 (0.53) & 84.54 (0.89)\\
			& EST         & 84.60 (0.76) & \textbf{85.78 (0.63)}\\
			\bottomrule[0.5pt]
		\end{tabular}
	} 
	\caption{\label{tab:results_ncaltech101} Object recognition accuracy (\%) of different deep nets with varying representations on N-Caltech101.}
\end{table}

\subsection{Results on N-Cars}
We then compare the results of EventDrop with the baselines on N-Cars. As can be seen from Table \ref{tab:results_ncars}, EventDrop outperforms the baselines with different deep learning architectures and event representations. The improvement on N-Cars dataset is generally greater than that on N-Caltech101 dataset. This might be because N-Cars is a real-world data dataset, and EventDrop works better with real-world cases where sensor noise and occlusion occur more likely than simulation (i.e., where the event camera looks at a projected scene rather than a real-world scene). The improvement of EventDrop can reach up to about 4.5\% (by ResNet-34 with EST representation). The Event Count and EST representations, which consider polarity information, perform better than the Event Frame and Voxel Grid that do not take polarity into account. The performance of the four deep nets is similar among the baselines, while ResNet-34 and MobileNet-V2 achieve better accuracy when the training data is augmented with EventDrop. 

\begin{table}
	\centering
	\resizebox{\linewidth}{!}{
		\begin{tabular}{cccccccc}
			\toprule[0.5pt]
			\multirow{2}{*}{\bf{Model}}  & \multirow{2}{*}{\bf{Representation}} &  \multicolumn{2}{c}{\textbf{Average Accuracy (Std)}} \\ \cline{3-4}
			& & \textbf{Baseline} & \textbf{EventDrop}  \\ 
			\midrule[0.5pt]
			\multirow{4}{*}{ResNet-34} 
			&  Event Frame & 91.83 (0.61) & 94.04 (0.19) \\
			& Event Count & 92.18 (0.34) & 95.20 (0.38)\\
			& Voxel Grid & 91.09 (0.46) & 93.61 (0.82)\\
			& EST & 91.03 (1.30) & \textbf{95.50 (0.18)}\\ \cline{1-4}
			
			\multirow{4}{*}{VGG-19} 
			& Event Frame & 91.61 (0.82) & 92.74 (0.81)\\
			& Event Count & 91.20 (0.53) & 93.19 (1.00)\\
			& Voxel Grid    & 91.45 (0.88) & \textbf{93.39 (0.86)}\\
			& EST         & 91.72 (0.85) & 93.12 (0.95)\\ \cline{1-4}

			\multirow{4}{*}{MobileNet-V2} 
			& Event Frame & 91.87 (0.54)& 93.64 (0.50)\\
			& Event Count & 92.70 (1.22)& \textbf{95.19 (0.71)}\\
			& Voxel Grid    & 91.18 (0.61)& 94.05 (0.38)\\
			& EST         & 91.71 (0.29)& 94.55 (0.45)\\ \cline{1-4}
			
			\multirow{4}{*}{Inception-V3} 
			& Event Frame  & 91.21 (0.56)  & 92.22 (2.73)\\
			& Event Count  & 91.16 (0.74) & 94.41 (0.99)\\
			& Voxel Grid     & 90.67 (0.96) & 92.12 (1.68)\\
			& EST          &  90.91 (1.78) &  \textbf{94.44 (0.72)}\\
			\bottomrule[0.5pt]
		\end{tabular}
	} 
	\caption{\label{tab:results_ncars} Object recognition accuracy (\%) of different deep nets with varying representations on N-Cars.}
\end{table}

\subsection{Comparison of Different Dropping Strategies}
We also compare the performance of different dropping strategies on both datasets. In the implementation of \textit{Drop by time}, \textit{Drop by area}, and \textit{Random drop} operations, the probability of each operation being conducted is set to 0.5, and the corresponding magnitude is randomly selected from the value set described in Section 4. For EventDrop, the three dropping strategies and \textit{Identity} operations are randomly selected to conduct with equal probability.
As demonstrated in Table \ref{tab:results_dropping_strategy}, EventDrop that integrates different dropping strategies outperforms the baselines on both datasets, and the improvement of EventDrop over the baselines is bigger on N-Cars dataset than on N-Caltech101 dataset. 

Specifically, for N-Caltech101 dataset, EventDrop and \textit{Drop by area} operations have better performance than \textit{Drop by time} and \textit{Random drop} operations in general. The \textit{Drop by time} operation seems not to improve the baselines when using Voxel Grid and EST representations but it improves the performance when using event frame and Event Count representations. This might be explained by N-Caltech101 being a simulated dataset in which the sensor noise and occlusion in time are negligible, and hence discarding some events that are selected randomly or triggered during a certain period of time does not increase the diversity of data.  
By contrast, for N-Cars dataset, all the dropping operations result in a better accuracy than the baselines. This might be because the real-world event dataset (N-Cars) suffers more from sensor noise and various occlusions, and dropping operations can better increase the data diversity.

\begin{table}  	
	\centering
	\resizebox{\linewidth}{!}{
		\begin{tabular}{cccccccc}
			\toprule[0.5pt]
			\multirow{2}{*}{\bf{Representation}} & \multirow{2}{*}{\bf{Dropping Strategy}} & \multicolumn{2}{c}{\textbf{Average Accuracy (Std)}} \\ \cline{3-4}
			& & \textbf{N-Caltech101} & \textbf{N-Cars}  \\ 
			\midrule[0.5pt]
			\multirow{5}{*}{Event Frame} 
			& Baseline & 77.39 (0.78) & 91.83 (0.61)\\ \cline{2-4}
			& Drop by time & 78.49 (0.70) & 92.81 (1.27) \\
			& Drop by area & 77.49 (0.71)& 92.59 (0.71)\\
			& Random drop & 77.19 (0.98) & 92.23 (0.30)\\ 
			& EventDrop & 78.20 (0.15) & \textbf{94.04 (0.19)}\\ \cline{1-4}
			
			\multirow{5}{*}{Event Count} 
			& Baseline & 77.75 (0.64) & 92.18 (0.34)\\ \cline{2-4}
			& Drop by time &  78.12 (0.83) & 93.91 (0.61)\\
			& Drop by area & 77.24 (0.80) & 93.81 (0.49)\\
			& Random drop &  77.68 (0.54)& 92.93 (0.87)\\ 
			& EventDrop & 78.30 (0.29)& \textbf{95.20 (0.38)} \\ \cline{1-4}
			
			\multirow{5}{*}{Voxel Grid} 
			& Baseline &  82.47 (0.80) & 91.09 (0.46)\\ \cline{2-4}
			& Drop by time & 80.80 (0.72)& 92.97 (0.44)\\
			& Drop by area & 83.84 (1.09)& 92.04 (0.66)\\
			& Random drop & 82.92 (1.00)&  91.29 (0.79)\\ 
			& EventDrop & 82.57 (0.42)& \textbf{93.61 (0.82)}\\ \cline{1-4}
			
			\multirow{5}{*}{EST}
			& Baseline & 83.91 (0.44) & 91.03 (1.30)\\ \cline{2-4}
			& Drop by time & 83.65 (0.59) & 94.73 (0.38)\\
			& Drop by area & 85.18 (0.83) & 92.71 (1.03)\\
			& Random drop &  84.07 (0.52) & 93.84 (0.52)\\ 
			& EventDrop & 85.15 (0.36) & \textbf{95.50 (0.18)} \\ 
			\bottomrule[0.5pt]
		\end{tabular}
	} 
	\caption{\label{tab:results_dropping_strategy} Accuracy (\%) comparison of different dropping strategies based on ResNet-34.}
\end{table}

\begin{figure}[!htb]
	\centering
	\begin{minipage}[t]{0.48\linewidth}
		\centering
		\includegraphics[width=\linewidth]{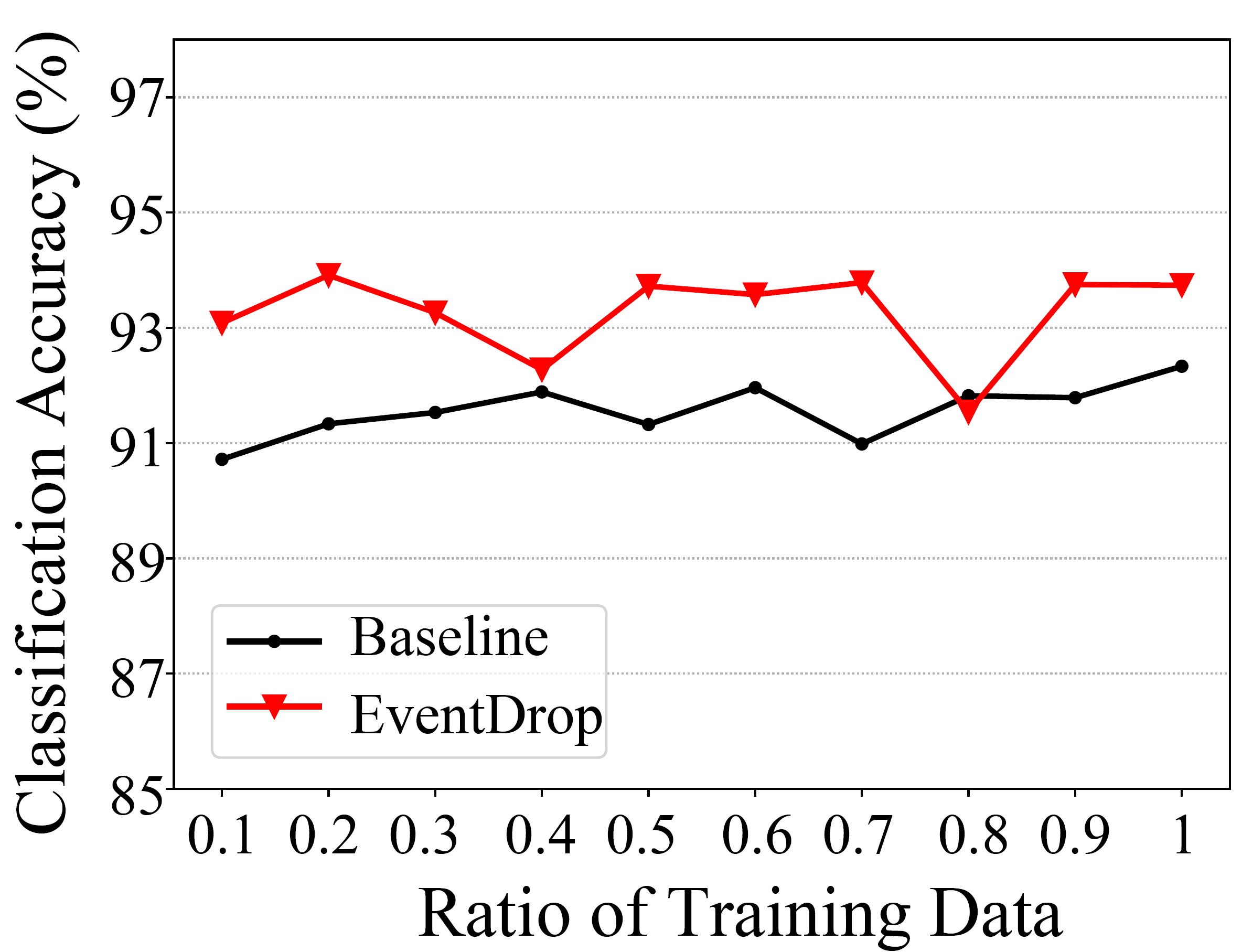} 
		\caption*{(a) Event frame}
	\end{minipage}
	~
	\begin{minipage}[t]{0.48\linewidth}
		\centering
		\includegraphics[width=\linewidth]{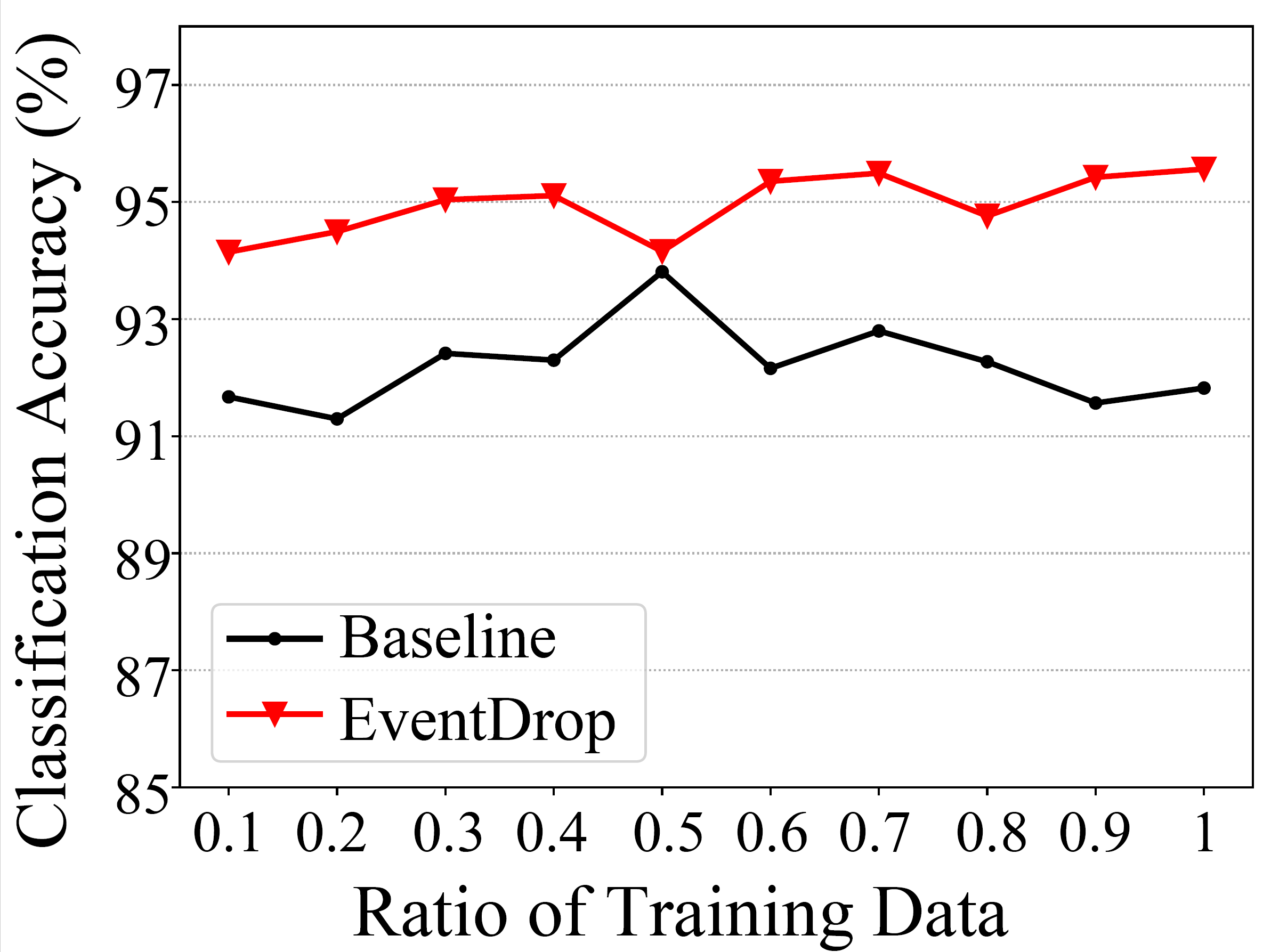} 
		\caption*{(b) Event Count}
	\end{minipage}
	~
	\begin{minipage}[t]{0.48\linewidth}
		\centering
		\includegraphics[width=\linewidth]{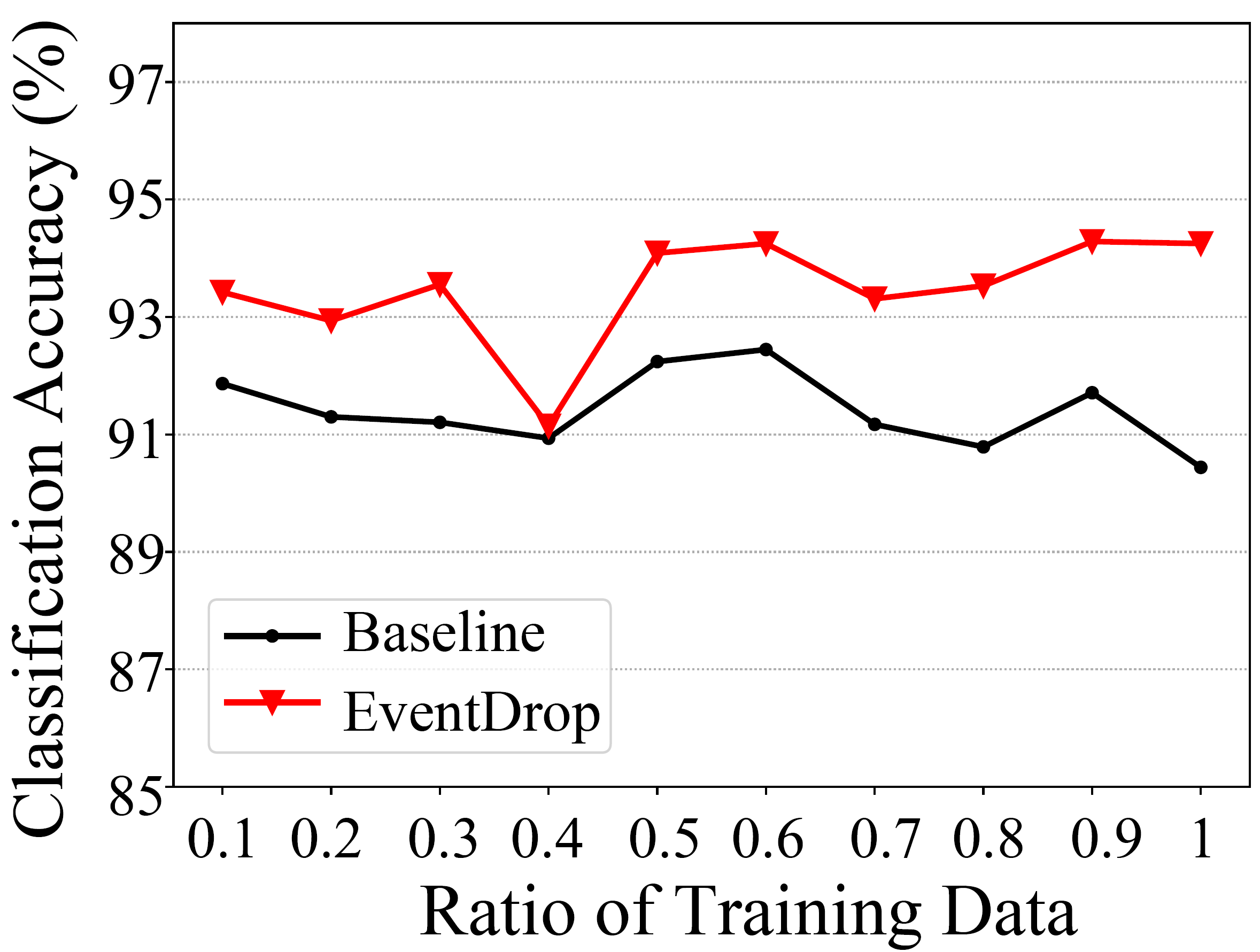} 
		\caption*{(c) Voxel Grid}
	\end{minipage}
	~
	\begin{minipage}[t]{0.48\linewidth}
		\centering
		\includegraphics[width=\linewidth]{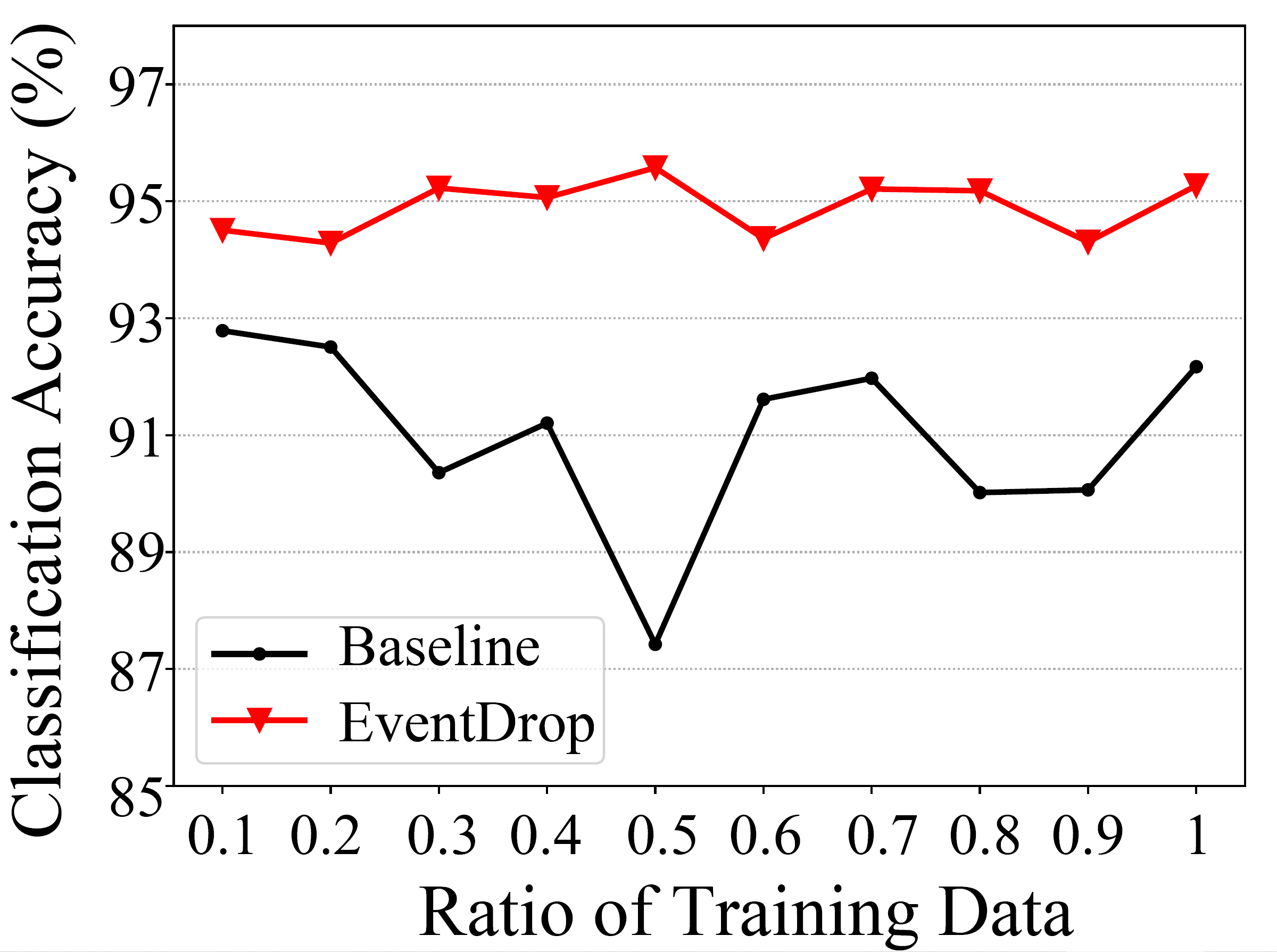} 
		\caption*{(d) EST}
	\end{minipage}
	\caption{Object classification accuracy using different ratios of training data on N-Cars with ResNet-34. }
	\label{fig:results_different_ratio}
\end{figure}

\subsection{Effect of the Amount of Training Data}
In this section, we analyze the effect of using different amounts of training data. The ratio we considered ranges from 0.1 to 1 where 0.1 represents only 10\% training data that are randomly selected are used to train the network. To reduce the search space, we fix the random seed that is shared by the baselines and EventDrop, and then compare their performance. Figure \ref{fig:results_different_ratio} shows that EventDrop consistently improves the baselines in general. It can achieve about 94\% accuracy even with only 10\% training data compared to about 91.5\% by the baseline. Although the improvement of EventDrop over the baseline is marginal when it is trained with certain ratios of training data, such a problem would obviously be reduced by averaging the results over more runs. It is also clear that the improvement of EventDrop is stable when using the more relatively robust EST representation. 

\section{Conclusion}
In this paper, we propose a new augmentation method for event-based learning, which we call EventDrop. It is easy to implement, computationally low-cost, and does not involve any parameter learning. 
We have demonstrated that by dropping events selected with 
certain strategies, we can significantly improve the object classification accuracy of different deep networks on two event datasets. While we show the application of our approach for event-based learning with deep nets, our approach can be also applied to learning with SNNs.
For future work, we will apply our approach to
other event-based learning tasks, such as visual inertial odometry, place recognition, traffic flow estimation, and simultaneous localization and mapping.

\bibliographystyle{named}
\bibliography{references}
\end{document}